\documentclass[11pt]{article}

\usepackage[preprint]{acl}

\usepackage{times}
\usepackage{latexsym}

\usepackage[T1]{fontenc}

\usepackage[utf8]{inputenc}

\usepackage{microtype}

\usepackage{inconsolata}

\usepackage{graphicx}
\usepackage{amsmath}
\usepackage{amsthm}
\usepackage{algorithm}
\usepackage{algorithmic}
\usepackage{amsfonts}

\usepackage{CJKutf8}

\usepackage{booktabs}
\usepackage{multirow}
\usepackage{colortbl}
\usepackage{adjustbox}
\usepackage{makecell}
\usepackage{nicematrix}
\usepackage{tikz}

\definecolor{s_grey}{RGB}{231,230,230}
\definecolor{s_cyan}{RGB}{204,229,230}
\definecolor{s_red}{RGB}{242,211,210}
\definecolor{s_green}{RGB}{209,217,211}
\definecolor{darkgreen}{RGB}{39,83,23}
\newcommand\modelname[0]{\texttt{SAS}}

\newcommand{\uptriangle}{
    \begin{tikzpicture}
        \fill[green!60!black] (0, 0) -- (-0.1, -0.2) -- (0.1, -0.2) -- cycle;
    \end{tikzpicture}
}

\newcommand{\downtriangle}{
    \begin{tikzpicture}
        \fill[red!80!black] (0, 0) -- (-0.1, 0.2) -- (0.1, 0.2) -- cycle;
    \end{tikzpicture}
}

\title{Can Large Language Models Resolve Semantic Discrepancy in Self-Destructive Subcultures? Evidence from Jirai Kei}

\author{
  Peng Wang$^{1,\,2,\,*}$~~
  Xilin Tao$^{1,}$\thanks{~~Equal Contribution.}~~
  Siyi Yao$^{3}$~~
  Jiageng Wu$^{2}$~~
  \\
  \textbf{Yuntao Zou$^{4}$~~
  Zhuotao Tian$^{5}$~~
  Libo Qin$^{5}$~~
  Dagang Li$^{1,}$}\thanks{~~Corresponding Author.}
  \\
	$^{1}$~School of Computer Science and Engineering, Macau University of Science and Technology, Macau, China \\
	$^{2}$~SKLPlanets, Macau University of Science and Technology, Macau, China\\
  $^{3}$~College of Software, Northeastern University, Shenyang, China\\
  $^{4}$~School of Energy and Power Engineering, Huazhong University of Science and Technology, Hubei, China\\
  $^{5}$~School of Computer Science and Technology, Harbin Institute of Technology (Shenzhen), Shenzhen, China\\
	\texttt{wpengxss@gmail.com}, \texttt{gaivrt@outlook.com}, \texttt{dgli@must.edu.mo}\\
}

\begin{document}
\maketitle
\begin{abstract}
Self-destructive behaviors are linked to complex psychological states and can be challenging to diagnose. These behaviors may be even harder to identify within subcultural groups due to their unique expressions. As large language models (LLMs) being deployed across various fields, some researchers have begun exploring their application for detecting self-destructive behaviors. Motivated by this, we investigate self-destructive behavior detection within subcultures using current LLM-based methods. However, these methods have two main challenges: (1) \textbf{\textit{Knowledge Lag}}: Subcultural slang evolves rapidly, faster than LLMs' training cycles; and (2) \textbf{\textit{Semantic Misalignment}}: it is challenging to grasp the specific and nuanced expressions unique to subcultures. To address these issues, we propose \textbf{\underline{S}}ubcultural \textbf{\underline{A}}lignment \textbf{\underline{S}}olver (\modelname{}), a multi-agent framework that incorporates automatic retrieval and subculture alignment, significantly boosting the performance of LLMs in detecting self-destructive behavior. Our experimental results show that \modelname{} outperforms the current advanced multi-agent framework OWL. Notably, it competes well with fine-tuned LLMs. We hope that \modelname{} will advance the field of self-destructive behavior detection in subcultural contexts and serve as a valuable resource for future researchers.
\end{abstract}


\section{Introduction}

\begin{quotation}
	\textit{``Everyone, deep in their hearts, is waiting for the end of the world to come.''}
	\begin{flushright}
		\par \textit{-- Haruki Murakami, 1Q84}
	\end{flushright}
\end{quotation}

\begin{figure}[t]
  \centering
  \includegraphics[width=\columnwidth]{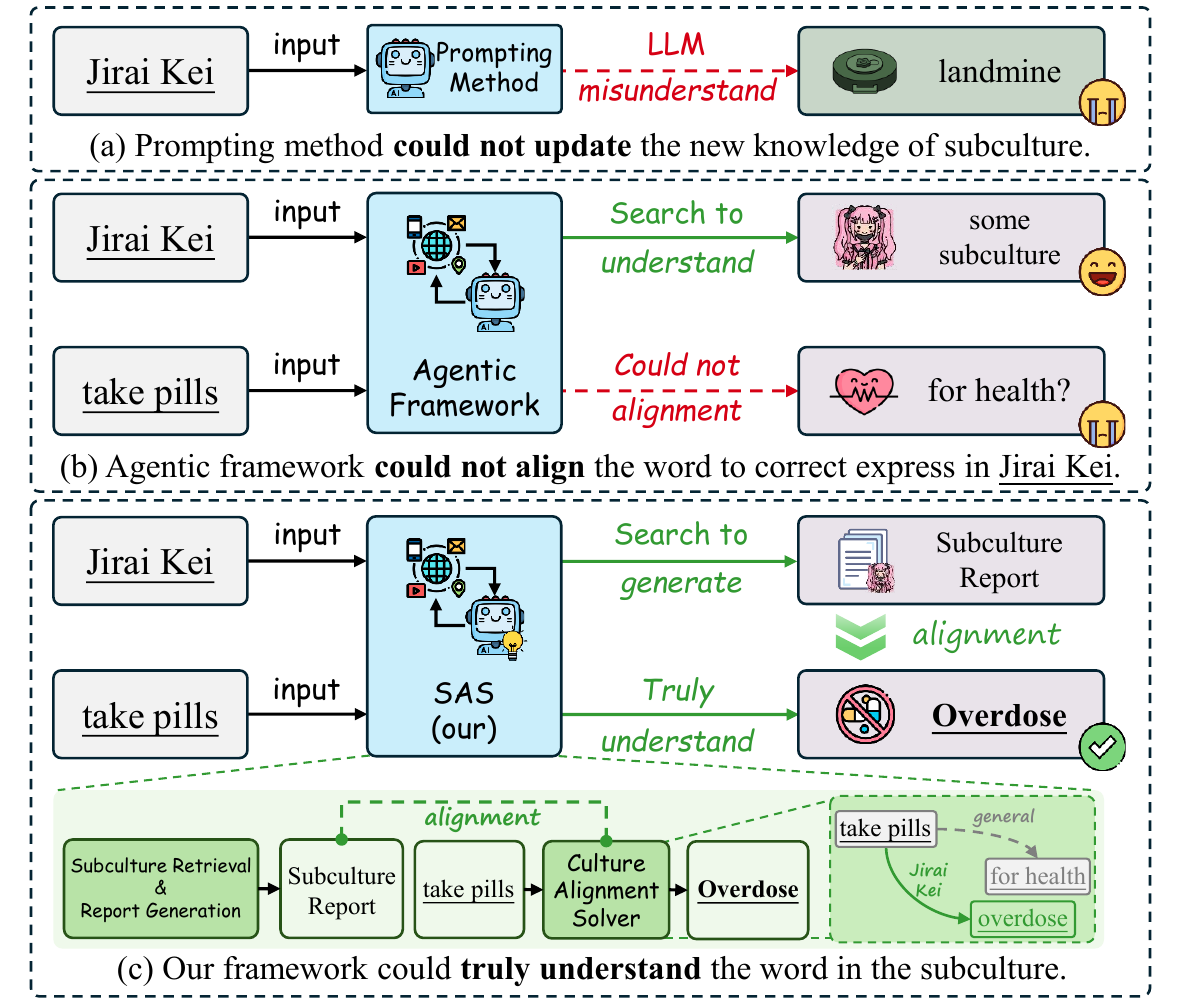}
  \caption{A comparative example of different methods. The prompting method often struggles to grasp emerging subcultures, which can result in misunderstandings by LLMs. In contrast, the agentic framework could understand Jirai Kei, but not the more detailed expressions. Our method could successfully summarize the subculture and effectively aligns expressions within this background. For instance, it recognizes that ``take pills'' in the context of Jirai Kei typically refers to an overdose.}
  \label{fig:intro}
\end{figure}

Self-destructive behaviors are actions that inflict direct or potential harm on oneself, frequently linked to complex psychological conditions~\cite{baumeister1988self,firestone1990suicide,van1991childhood,baumeister1997esteem}. Especially with the development of social media, people are increasingly inclined to express their feelings online. While online platforms facilitate connection, they also foster ``echo chambers'' where harmful content is validated rather than mitigated~\cite{cinelli2021echo}. A critical challenge in these environments is the emergence of niche subcultures; alternative subcultures may be particularly vulnerable to self-destructive behavior~\cite{rutledge2008vulnerable}. These communities develop unique linguistic norms and evolving slang to conceal harmful intents or signal group identity.

To detect such behaviors, prior research has largely focused on general mental health modeling. For instance, \citet{stade2024large} advocate for the potential of large language models (LLMs) to enhance clinical psychotherapy within broader behavioral healthcare. \citet{sathvik-etal-2025-help} introduce M-HELP, a novel dataset to identify help-seeking signals on social media. \citet{li2025can} reveal that current LLMs often struggle to detect implicit suicidal ideation, indicating a gap in understanding nuanced expressions between common and special meaning. Recently, specific benchmarks like JiraiBench~\cite{xiao2025jiraibench}, a dataset inspired by the landmine subculture, aimed at aiding the detection of self-destructive behaviors within subcultural settings. However, a critical gap remains: existing methods, whether specialized small models or general-purpose LLMs, operate on static knowledge bases. They frequently fail to grasp the rapidly evolving terminology and the ``semantic gap'' present in subcultures, where seemingly innocuous words may carry severe self-destructive implications. This naturally raises a question:

\textit{Is LLM-based method adequate for understanding the self-destructive subculture of the Internet?}

To address this question, we evaluate LLM-based methods in the context of the emerging Jirai Kei\footnote{\url{https://aesthetics.fandom.com/wiki/Jirai_Kei}}. Literally translating to ``landmine type,'' this subculture often disguises severe self-destructive tendencies behind a sick-cute aesthetic and cryptic slang, rendering harmful intent opaque to models.
Our empirical evaluation yields a \underline{\textbf{negative}} answer. Our experimental results show that: (1) \textbf{Knowledge Lag}: current methods struggle to enable large language models to understand new terminology within subcultures. Subcultural slang evolves faster than model training cycles, so LLMs cannot update relevant knowledge on time, as shown in Figure~\ref{fig:intro} (a). (2) \textbf{Semantic Misalignment}: While advanced multi-agent frameworks can automatically utilize multi-view information or retrieval tools, they do not effectively bridge the understanding gap between different subcultures. The semantic gap between general usage and subcultural intent leads to misunderstanding, as shown in Figure~\ref{fig:intro} (b).

This suggests that to improve the comprehension capabilities of large language models in the context of self-destructive subcultures, a combination of \textbf{knowledge updates} and \textbf{subculture alignment} is necessary.
To address this, we propose \textbf{\underline{S}}ubcultural \textbf{\underline{A}}lignment \textbf{\underline{S}}olver (\modelname{}), which could enhance the understanding of the subculture in LLMs. \modelname{} consists of three parts: ($i$) \textbf{\textit{Subculture Retrieval}}: Given a target subculture, it automatically searches the internet for relevant knowledge, returning valid results related to that subculture. ($ii$) \textbf{\textit{Alignment Report Generation}}: Based on the retrieval results, it generates an analysis report that highlights the background and unique terminology associated with the target subculture. ($iii$) \textbf{\textit{Culture Alignment Solver}}: Utilizing the subculture alignment report, it first comprehends and interprets the input content, then addresses the task based on this understanding. Finally, \modelname{} can significantly improve the ability of LLMs to identify harmful subcultural content, which is shown in Figure~\ref{fig:intro} (c).

We conduct experiments using the subculture benchmark JiraiBench~\cite{xiao2025jiraibench}, and the results indicate that \modelname{} outperforms other methods, even the advanced general agentic framework OWL~\cite{hu2026owl}. Notably, our framework competes effectively with fine-tuned LLMs in terms of performance. Additionally, we test \modelname{} on various other subcultures, further demonstrating its capability to identify harmful subcultural content, even for large language models that lack prior exposure to relevant information. We hope this work will serve as a general diagnostic framework that can adapt to rapidly evolving cultural developments, enhancing the efficiency of identifying self-destructive behaviors in subcultural contexts.

Our contributions can be summarized as follows:
\begin{itemize}
  \item To the best of our knowledge, we are the first to consider the subculture alignment within LLMs. We evaluate the performance of various methods, and it indicates that current methods struggle with subcultures.
  \item We propose the Subcultural Alignment Solver (\modelname{}), which automatically retrieves and summarizes subcultures. \modelname{} effectively enhances the understanding of LLMs and allows them to solve tasks more accurately.
  \item Final experimental results demonstrate the effectiveness of our framework. Importantly, \modelname{} can compete with fine-tuned LLMs and generalize to other subcultures.
\end{itemize}

\begin{figure*}[t]
  \centering
  \includegraphics[width=1.00\textwidth]{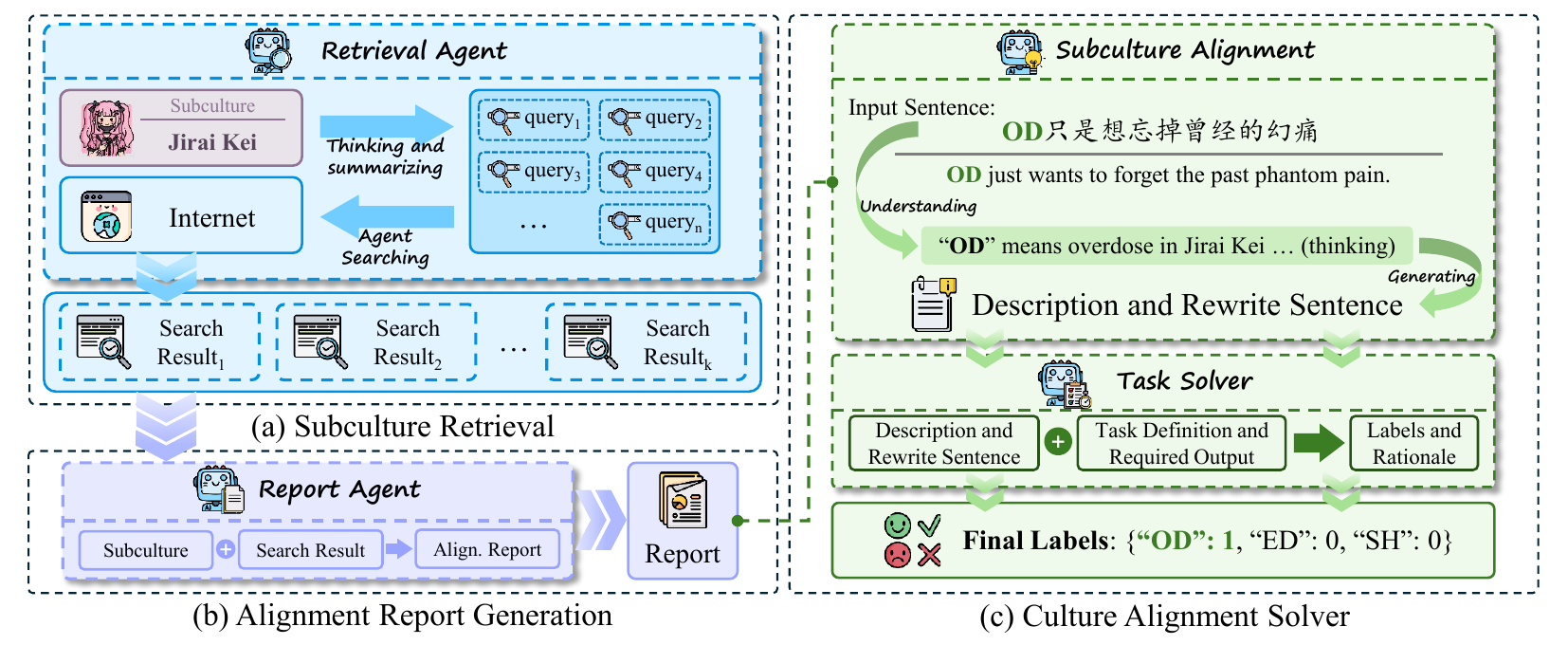}
  \caption{The main framework of \modelname{}. In \modelname{}, Subculture Retrieval retrieves semantically relevant information based on the target subculture. Alignment Report Generation creates a comprehensive subculture report based on these results. In the Culture Alignment Solver, the input sentence is identified and interpreted in relation to the report; it then outputs the final labels based on the interpreted content. The example in (c) is the actual input sentence from JiraiBench~\cite{xiao2025jiraibench}, which we have translated into English for demonstration purposes.}
  \label{fig:main}
\end{figure*}

\section{Background}
Jirai Kei, translating to ``landmine type'', is a youthful and trendy subculture that originated in Japan. It took on a negative connotation, referring to somebody as emotionally unstable and prone to breakdowns. This subculture has been linked to troubling events like self-harm and overdoses, prompting some teenagers to engage in self-destructive behaviors in an effort to gain acceptance within it.

To evaluate the models' ability of detecting self-destructive behaviors on social media in LLMs, researchers introduce JiraiBench~\cite{xiao2025jiraibench}, a benchmark specifically focused on the Jirai Kei subculture. In JiraiBench, the self-destructive behavior detection can be defined as a label classification task. Given an input sentence from social media, the model should assess whether the sentence expresses any of the following behaviors: Drug Overdose (OD), Eating Disorders (ED), or Self-Harming (SH). For each behavior, the model outputs a classification label: Non-concerning (0), First-person (1), or Third-party (2).


\section{Subcultural Alignment Solver (\modelname{})}

\modelname{} has three part: ($i$) Subculture Retrieval (\S~\ref{subsec:subculture_retrieval}), ($ii$) Alignment Report Generation (\S~\ref{subsec:alignment_report_generation}), and ($iii$) Culture Alignment Solver (\S~\ref{subsec:culture_alignment_solver}). The main framework are illustrated in Figure~\ref{fig:main}.

\subsection{Subculture Retrieval}
\label{subsec:subculture_retrieval}
While large language models may not have internal knowledge of subcultures, the presence of these subcultures on the internet leaves discernible traces. This allows for the understanding of the subculture through the retrieval of relevant information. Given target subculture $\mathcal{S}$, target language $\mathcal{L}$, and search country $\mathcal{C}$. The retrieval agent involves a process that searches for information based on $\mathcal{S}, \mathcal{L}, \mathcal{C}$ and returns $k$ results from text searches.

Firstly, the retrieval agent will consider how to retrieve information based on the input subculture and provide a suitable set of search queries $\mathbb{Q}_n=\left\{q_1, q_2, ..., q_n\right\}$. Then the retrieval agent searches the internet for relevant content using the provided queries $\mathbb{Q}_n$. It returns $m$ results for each query $q_i, i\in \left\{1, 2, ..., n\right\}$. Finally, the retrieval agent returns $k$ results, $k=m*n$.
This process can be defined as:
\begin{equation}
  \mathbb{R}_{k}=\operatorname{Retrieval\,\,Agent}\left(\mathcal{S}, \mathcal{L}, \mathcal{C}\right),
\end{equation}
which $\mathbb{R}_{k}=\left\{r_1, r_2, ..., r_k\right\}$ presents the search results. Unlike standard retrieval-augmented generation, where search queries are typically formed without considering cultural context, our retrieval agent explicitly grounds its queries in the target subculture. This helps the retrieved content better capture subculture-specific semantics, facilitating more effective alignment in subsequent stages.

\subsection{Alignment Report Generation}
\label{subsec:alignment_report_generation}
After obtaining the search results $\mathbb{R}_{k}$, the report agent will create a reliable subculture background report that incorporates all relevant information. This report include essential elements such as an introduction, background information, and specific terminology. This process can be defined as:
\begin{equation}
  \mathcal{R}_{A}=\underset{ARG}{\operatorname{argmax}} \,\mathrm{P}\left(\mathcal{R}_{A}^{n}|\mathcal{P}_{ARG}, \mathbb{R}_{k}\right),
\end{equation}
where $\mathcal{P}_{ARG}$ represents the prompting about report generation. $\mathcal{R}_{A}^{n}$ presents potential reasoning path by inputting $\mathcal{P}_{ARG}$ and $\mathbb{R}_{k}$. $\mathcal{R}_{A}$ presents the alignment report.

\subsection{Culture Alignment Solver}
\label{subsec:culture_alignment_solver}
Culture Alignment Solver involves two parts: ($i$) Subculture Alignment and ($ii$) Task Solver.

First, Subculture Alignment aligns the representations of subcultures based on both the report $\mathcal{R}_{A}$ and the input sentence $\mathcal{I}$. Specifically, it interprets the input sentence through the lens of the subculture report, identifying subculture-specific expressions and generating a description $\mathcal{D}_{e}$ of their intended meanings, along with a rewritten sentence $\mathcal{D}_{r}$ that clarifies these expressions in plain language. $\mathcal{D}_{e}$ is a set of term-explanation pairs, where each pair consists of a subculture term detected in the input and its explanation explicitly grounded in the reference report. This process can be defined as:
\begin{equation}
  \{\mathcal{D}_{e},\mathcal{D}_{r}\}=\underset{SA}{\operatorname{argmax}} \,\mathrm{P}\left(\{\mathcal{D}_{e},\mathcal{D}_{r}\}^{n}|\mathcal{P}_{SA}, \mathcal{I}, \mathcal{R}_{A}\right),
\end{equation}
where $\mathcal{P}_{SA}$ presents the prompting about subculture alignment. $\{\mathcal{D}_{e},\mathcal{D}_{r}\}^{n}$ represents potential reasoning path by inputting $\mathcal{P}_{SA}$, $\mathcal{I}$, and $\mathcal{R}_{A}$.

Then, building on the understanding of subculture alignment, the Task Solver produces the final answer based on the specified task $\mathcal{T}$. For our work, the Task Solver generates final labels $\mathbb{L}=\left\{l_{OD}, l_{ED}, l_{SH}\right\}$ including OD, ED, and SH. It can be defined as:
\begin{equation}
  \mathbb{L}=\underset{TS}{\operatorname{argmax}} \,\mathrm{P}\left(\mathbb{L}^{n}|\mathcal{P}_{TS}, \mathcal{T}, \mathcal{H}_{SA}\right),
\end{equation}
where $\mathcal{H}_{SA}$ is the dialogue history in Subculture Alignment, including $\mathcal{D}_{e}$ and $\mathcal{D}_{r}$. $\mathbb{L}^{n}$ presents potential reasoning path.

\section{Experiments}

\subsection{Experimental Settings}
Following previous work~\cite{xiao2025jiraibench}, we use JiraiBench for experiments, which is the self-destructive behavior detection task in the subculture Jirai Kei. We use the macro F1 score in OD, ED, and SH as the evaluation metrics.

We conduct experiments on five large language models: Qwen-2.5-7B~\cite{qwen2025qwen25technicalreport}, Llama-3.1-8B~\cite{grattafiori2024llama}, DeepSeek V3.2~\cite{liu2025deepseek}, Ministral-3-8B~\cite{ministral-3-8b}, and Gemma-3-12B-it~\cite{team2025gemma}. All models are in a non-thinking mode. For all LLMs, the temperature is 0.0-0.3, and the top-p is 1. We use the same model throughout the entire \modelname{}. For all baselines, we conduct a single run to evaluate performance. For \modelname{}, we conduct three repeated experiments and report the mean $\pm$ standard deviation.

\begin{table*}[ht]
    \centering
    \begin{adjustbox}{width=0.975\textwidth}
        \begin{tabular}{ccccccc}
    \toprule
    Method & Task  & Qwen-2.5-7B & Llama-3.1-8B & DeepSeek V3.2 & Ministral-3-8B & Gemma-3-12B-it \\
    \midrule
    \multicolumn{1}{c}{\multirow{3}[2]{*}{\makecell[c]{Zero-shot\\\colorbox{s_red}{EACL}}}} & OD    & 0.5206  & 0.3622  & 0.7325  & \cellcolor[rgb]{ 1,  .949,  .8}0.6314  & 0.5150  \\
          & ED    & \cellcolor[rgb]{ .988,  .894,  .839}0.4452 & \cellcolor[rgb]{ 1,  .949,  .8}0.5295 & 0.7415  & \cellcolor[rgb]{ 1,  .949,  .8}0.5758  & 0.4629  \\
          & SH    & 0.3971  & \cellcolor[rgb]{ 1,  .949,  .8}0.3900 & 0.5733  & 0.4833  & 0.4592  \\
    \midrule
    \multicolumn{1}{c}{\multirow{3}[2]{*}{\makecell[c]{Zero-shot CoT\\\colorbox{s_cyan}{NeurIPS}}}} & OD    & 0.4759  & 0.3308  & 0.7235  & \cellcolor[rgb]{ .886,  .937,  .855}0.6338  & 0.5090  \\
          & ED    & 0.5242  & 0.5256  & 0.7386  & 0.5753  & 0.4746  \\
          & SH    & \cellcolor[rgb]{ 1,  .949,  .8}0.4999 & 0.3629  & 0.5702  & 0.4886  & 0.4705  \\
    \midrule
    \multicolumn{1}{c}{\multirow{3}[2]{*}{\makecell[c]{Plan-and-Solve\\\colorbox{s_red}{ACL}}}} & OD    & \cellcolor[rgb]{ .988,  .894,  .839}0.4730 & \cellcolor[rgb]{ .988,  .894,  .839}0.2842 & 0.7025  & 0.6294  & \cellcolor[rgb]{ .988,  .894,  .839}0.5082 \\
          & ED    & 0.5317  & 0.4523  & \cellcolor[rgb]{ .988,  .894,  .839}0.7274 & 0.5396  & 0.4556  \\
          & SH    & 0.4992  & \cellcolor[rgb]{ .988,  .894,  .839}0.3274 & \cellcolor[rgb]{ .988,  .894,  .839}0.5591 & 0.4777  & 0.4594  \\
    \midrule
    \multicolumn{1}{c}{\multirow{3}[1]{*}{\makecell[c]{Self-Refine\\\colorbox{s_cyan}{NeurIPS}}}} & OD    & \cellcolor[rgb]{ .886,  .937,  .855}0.5785 & \cellcolor[rgb]{ 1,  .949,  .8}0.4221 & \cellcolor[rgb]{ 1,  .949,  .8}0.7652 & \cellcolor[rgb]{ .988,  .894,  .839}0.5483  & \cellcolor[rgb]{ .886,  .937,  .855}0.6549 \\
          & ED    & \cellcolor[rgb]{ .886,  .937,  .855}0.6136 & 0.4976  & \cellcolor[rgb]{ 1,  .949,  .8}0.7694 & 0.5280  & 0.6866  \\
          & SH    & \cellcolor[rgb]{ .988,  .894,  .839}0.3480 & 0.3883  & 0.5792  & \cellcolor[rgb]{ 1,  .949,  .8}0.4930  & \cellcolor[rgb]{ .886,  .937,  .855}0.5333 \\
      \midrule
    \multicolumn{1}{c}{\multirow{3}[1]{*}{\makecell[c]{S$^3$ Agent\\\colorbox{s_green}{ACM TOMM}}}} & OD    & 0.5084  & 0.3463  & \cellcolor[rgb]{ .886,  .937,  .855}0.7743 & 0.5727  & \cellcolor[rgb]{ 1,  .949,  .8}0.5769 \\
          & ED    & \cellcolor[rgb]{ 1,  .949,  .8}0.5524 & 0.3956  & 0.7585  & 0.5299  & \cellcolor[rgb]{ 1,  .949,  .8}0.6906 \\
          & SH    & 0.4718  & 0.3472  & \cellcolor[rgb]{ 1,  .949,  .8}0.6180 & \cellcolor[rgb]{ .988,  .894,  .839}0.4582  & 0.4603  \\
    \midrule
    \multicolumn{1}{c}{\multirow{3}[2]{*}{\makecell[c]{OWL\\\colorbox{s_cyan}{NeurIPS}}}} & OD    & 0.5015  & 0.3457  & \cellcolor[rgb]{ .988,  .894,  .839}0.6397 & 0.5855  & 0.5530  \\
          & ED    & 0.5290  & \cellcolor[rgb]{ .988,  .894,  .839}0.3879 & 0.7686  & \cellcolor[rgb]{ .988,  .894,  .839}0.5018  & \cellcolor[rgb]{ .988,  .894,  .839}0.4072 \\
          & SH    & 0.4703  & 0.3583  & 0.5975  & 0.4657  & \cellcolor[rgb]{ .988,  .894,  .839}0.4312 \\
    \midrule
    \multirow{3}[2]{*}{\texttt{SAS} (English version)} & OD    & \cellcolor[rgb]{ 1,  .949,  .8}0.5600 $\pm$ 0.0202 & \cellcolor[rgb]{ .886,  .937,  .855}0.4288 $\pm$ 0.0371 & 0.7538 $\pm$ 0.0046 & \cellcolor[rgb]{ .988,  .894,  .839}0.4645 $\pm$ 0.0067 & 0.5403 $\pm$ 0.0103 \\
          & ED    & \cellcolor[rgb]{ 1,  .949,  .8}0.5859 $\pm$ 0.0129 & \cellcolor[rgb]{ .886,  .937,  .855}0.6102 $\pm$ 0.0156 & \cellcolor[rgb]{ .886,  .937,  .855}0.8080 $\pm$ 0.0043 & \cellcolor[rgb]{ .886,  .937,  .855}0.7302 $\pm$ 0.0065 & \cellcolor[rgb]{ .886,  .937,  .855}0.7328 $\pm$ 0.0043 \\
          & SH    & \cellcolor[rgb]{ .886,  .937,  .855}0.5541 $\pm$ 0.0181 & \cellcolor[rgb]{ .886,  .937,  .855}0.4665 $\pm$ 0.0196 & \cellcolor[rgb]{ .886,  .937,  .855}0.6675 $\pm$ 0.0022 & \cellcolor[rgb]{ .886,  .937,  .855}0.5241 $\pm$ 0.0073 & \cellcolor[rgb]{ 1,  .949,  .8}0.5134 $\pm$ 0.0075 \\
    \bottomrule
    \end{tabular}%
    \end{adjustbox}
    \caption{Main results of baselines and our method. The \colorbox[rgb]{ .886,  .937,  .855}{green}, \colorbox[rgb]{ 1,  .949,  .8}{yellow} and \colorbox[rgb]{ .988,  .894,  .839}{red} respectively represents the best, the second best, and the worst performance of task achieved by this method on the current model. The publication is indicated under the method name.}
    \label{tab:main}%
\end{table*}%

\subsection{Baselines}
We evaluate six methods: (1) prompting methods involves zero-shot~\cite{xiao2025jiraibench}, zero-shot Chain-of-Thought (CoT)~\cite{kojima2022large}, and Plan-and-Solve~\cite{wang-etal-2023-plan}; (2) multi-agent frameworks includes Self-Refine~\cite{madaan2023self}, S$^3$ Agent~\cite{10.1145/3690642}, and Optimized Workforce Learning (OWL)~\cite{hu2026owl}. For all methods, we use English as the prompting language.
Detailed description of these baselines are following as:

\begin{itemize}
  \item Zero-shot~\cite{xiao2025jiraibench}: Based on the zero-shot prompting officially implemented by JiraiBench, it uses three different prompts to obtain the label for the input sentence.
  \item Zero-shot CoT~\cite{kojima2022large}: Based on zero-shot prompting, zero-shot CoT improves the performance of LLMs by using ``Let's think step by step!''.
  \item Plan-and-Solve~\cite{wang-etal-2023-plan}: Plan-and-Solve enhances LLMs by using ``Let's first understand the problem and devise a plan to solve the problem. Then, let's carry out the plan to solve the problem step by step''.
  \item Self-Refine~\cite{madaan2023self}: Self-Refine utilizes self-feedback of LLMs to continuously optimize the output and improve performance. In our experiments, we set the feedback limit to three to save costs.
  \item S$^3$ Agent~\cite{10.1145/3690642}: S$^3$ Agent uses three views: superficial expression, sentiment information, and semantic expression to enhance the ability of LLMs in sarcasm feelings.
  \item OWL~\cite{hu2026owl}: OWL is a hierarchical multi-agent framework that can autonomously add or modify worker agents and select tools in the tools pool based on the task, which makes OWL highly generalizable.
\end{itemize}


\subsection{Main Results}
The main results are shown in Table~\ref{tab:main}. Based on these results, we can obtain the following findings:

\textbf{\textit{(1) Prompting methods struggle to handle subculture-related tasks.}}
Prompting methods struggle to effectively detect self-destructive behaviors within subcultures, as evidenced by the fact that most prompting methods are less effective than other frameworks. Compared to zero-shot, the advantage of incorporating additional prompts is minimal. This indicates that existing prompting methods are insufficient for fostering a deeper understanding of tasks specific to these subcultures.

\textbf{\textit{(2) The agentic framework can enhance the ability of LLMs to comprehend subcultures.}}
Compared to prompting methods, the agentic framework demonstrates some improvement in this task. Surprisingly, self-refine proves effective across multiple LLMs, indicating that reflective processes can enhance LLMs' grasp of psychology-related tasks. S$^3$ Agent also exhibited some improvement, but the gains were not substantial.

\textbf{\textit{(3) Addressing the unique expressions within subcultures solely through the search tool is challenging.}}
In the advanced multi-agent framework OWL, the capability to autonomously plan tools for general tasks does not result in strong performance in detecting self-destructive behavior, as it falls short compared to other methods on most metrics. This indicates that, despite the OWL's ability to autonomously retrieve information, its performance remains limited due to a lack of understanding of subcultures. The agentic framework must grasp the deeper meanings behind specific expressions to make more accurate judgments within subcultures.

\textbf{\textit{(4) Our framework achieves advanced performance of detecting self-destructive behaviors in subcultures.}}
Compared to these baselines, our method achieves the state-of-the-art performance across most models, especially enhancing smaller LLMs. This suggests that our framework can enhance model performance on subculture tasks through retrieval and alignment. Notably, our framework demonstrates competitive zero-shot performance when compared to fine-tuned models in Qwen-2.5-7B, further underscoring the effectiveness of \modelname{}.

\begin{table}[ht]
    \centering
    \begin{adjustbox}{width=1.000\columnwidth}
    \begin{tabular}{ccccc}
    \toprule
    Method & Task  & Qwen-2.5-7B & Llama-3.1-8B & DeepSeek V3.2 \\
    \midrule
    \multirow{3}[2]{*}{\texttt{SAS}} & OD    & 0.5600 $\pm$ 0.0202 & 0.4288 $\pm$ 0.0371 & 0.7538 $\pm$ 0.0046 \\
          & ED    & 0.5859 $\pm$ 0.0129 & 0.6102 $\pm$ 0.0156 & 0.8080 $\pm$ 0.0043 \\
          & SH    & 0.5541 $\pm$ 0.0181 & 0.4665 $\pm$ 0.0196 & 0.6675 $\pm$ 0.0022 \\
    \midrule
    \multirow{3}[2]{*}{w/o CAS} & OD    & 0.5676 \uptriangle & 0.4580 \uptriangle & 0.7962 \uptriangle \\
          & ED    & 0.5613 \downtriangle & 0.5045 \downtriangle & 0.8002 \downtriangle \\
          & SH    & 0.4648 \downtriangle & 0.4497 \downtriangle & 0.6310 \downtriangle \\
    \bottomrule
    \end{tabular}%
    \end{adjustbox}
    \caption{Ablation study. The w/o CAS presents the \modelname{} without Culture Alignment Solver. \uptriangle represents a decline in performance, \downtriangle represents an improvement in performance.}
    \label{tab:abl}%
\end{table}%

\subsection{Ablation Study}

To investigate the effectiveness of the components in \modelname{}, we conduct ablation experiments. Specifically, we replace the Culture Alignment Solver in \modelname{} with a regular Solver, while keeping the retrieval information input, to explore the performance without culture alignment.

The final results are shown in Table~\ref{tab:abl}. As can be seen, after removing the Culture Alignment Solver, the performance of the three LLMs on the other subtasks, except for OD, dropped significantly. This demonstrates the effectiveness of culture alignment in \modelname{} for understanding subcultures in LLMs.

\section{Analysis}


\begin{figure}[t]
  \centering
  \includegraphics[width=0.9\columnwidth]{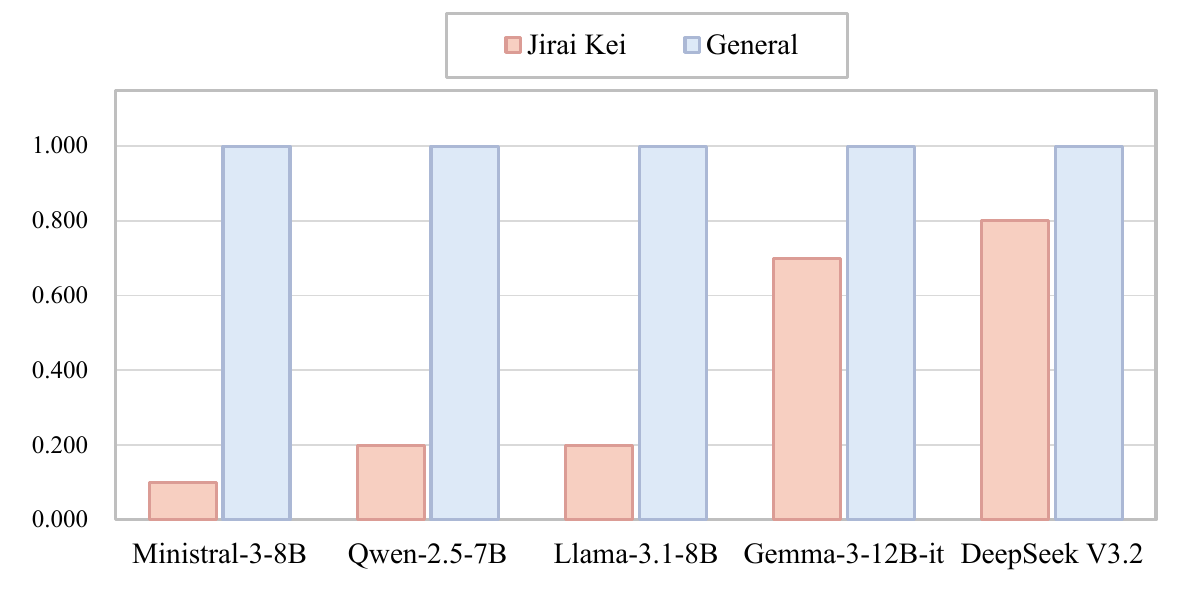}
  \caption{Performance of LLMs in Jirai Kei and General knowledge.}
  \label{fig:understand_1}
\end{figure}

\begin{figure}[t]
  \centering
  \includegraphics[width=0.9\columnwidth]{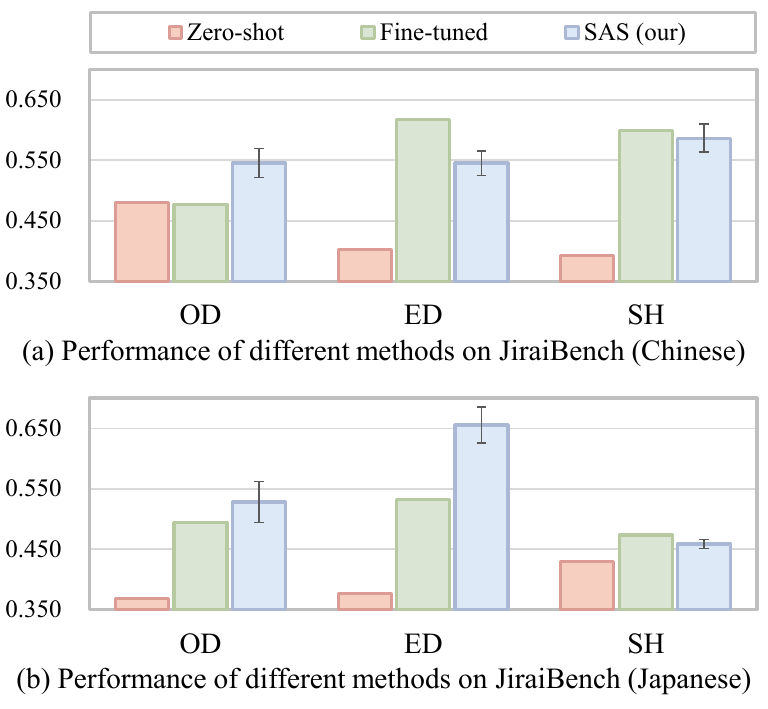}
  \caption{Performance of Qwen-2.5-7B under different methods, with the fine-tuned performance data sourced from Jirai-Qwen~\cite{xiao2025jiraibench}. In \modelname{}, we report mean $\pm$ SD ($n=3$).}
  \label{fig:finetune}
\end{figure}

\begin{figure}[t]
  \centering
  \includegraphics[width=1.0\columnwidth]{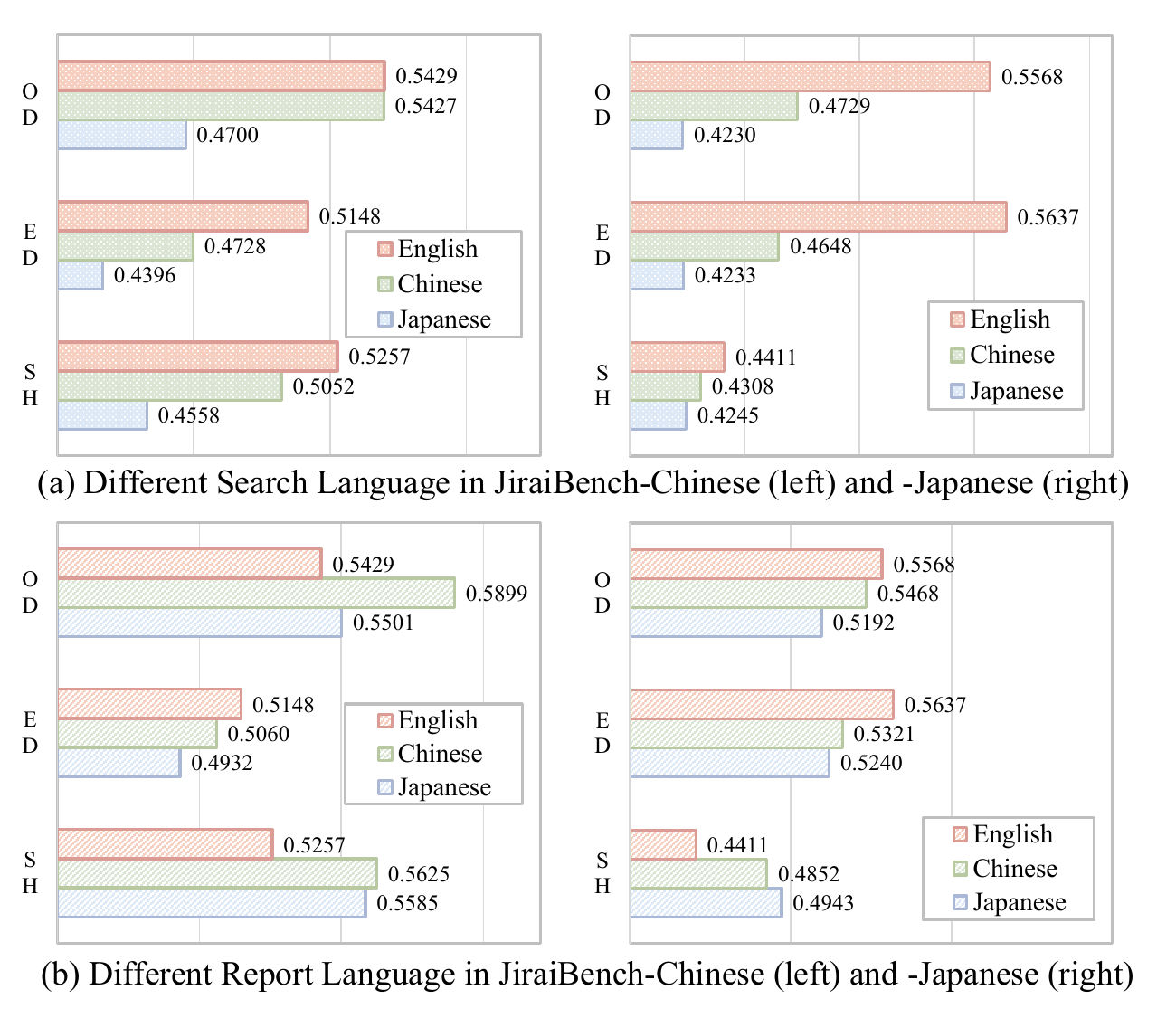}
  \caption{Performance comparison across multilingual retrieval and reporting. The prompt language for each trial is aligned with the respective local language. To ensure comparability, non-English reports were generated via controlled translation from the English report.}
  \label{fig:multilingual}
\end{figure}

\subsection{LLMs that have a better understanding of subcultures generally perform better}
\label{subsec:understanding_1}
To better investigate the performance differences among various LLMs on Jirai Kei, we create a set of 20 question-answer pairs that include both Jirai Kei-related content and general knowledge, following the SimpleQA~\cite{wei2024measuring}. This allows us to assess the knowledge capabilities of LLMs across different domains. The specific questions can be found in Appendix~\ref{appendix:qa-pairs}. All final answers were evaluated manually, and the results are presented in Figure~\ref{fig:understand_1}.

The findings indicate that LLM performance on JiraiBench correlates positively with their knowledge capabilities. This suggests that LLMs with more expertise in subcultural topics possess a distinct advantage in this task, as they can fully comprehend subcultural terminology and make accurate judgments. Moreover, within our framework, Qwen-2.5-7B and Llama-3.1-8B can achieve performance levels surpassing the original performance of Gemma-3-12B-it. This suggests that our framework effectively helps LLMs lacking subcultural knowledge to better understand context, ultimately improving their overall performance.

\subsection{Fine-tuning is effective, but it doesn't provide sufficient improvement}
\label{subsec:finetune}
To compare the performance of the fine-tuned model, we evaluate three methods of Qwen-2.5-7B: zero-shot, fine-tuned, and \modelname{}, all using English prompts. The final results are shown in Figure~\ref{fig:finetune}.

The results indicate that fine-tuning leads to significant performance improvements over the zero-shot approach, with an average gain of 11\% in Chinese and 28\% in Japanese. However, the fine-tuned model is nearly ineffective when applied to other subcultural domains. In contrast, our framework \modelname{}, which does not require fine-tuning, performs well on most metrics when compared to the fine-tuned model. This further highlights the effectiveness of \modelname{} for subcultural alignment.

\subsection{English dominates retrieval, while reporting benefits from lingual adaptation}
\label{subsec:multilingual}
To systematically evaluate the impact of search and reporting languages on \modelname{}, we conduct a comparative analysis using Qwen-2.5-7B. The results of this experiment are obtained using a single run. The quantitative results are illustrated in Figure~\ref{fig:multilingual}.

The findings reveal that performance with English search exceeds that of the other two languages. We attribute this discrepancy to differences in retrieval results associated with language switching, English retrieval typically provides richer content than other languages. Interestingly, while English dominates the retrieval phase, it does not universally excel as the optimal reporting language in cross-lingual contexts. Moving forward, combining robust retrieval results with the language that fits the subculture will be a promising direction.

\begin{figure}[t]
  \centering
  \includegraphics[width=1.000\columnwidth]{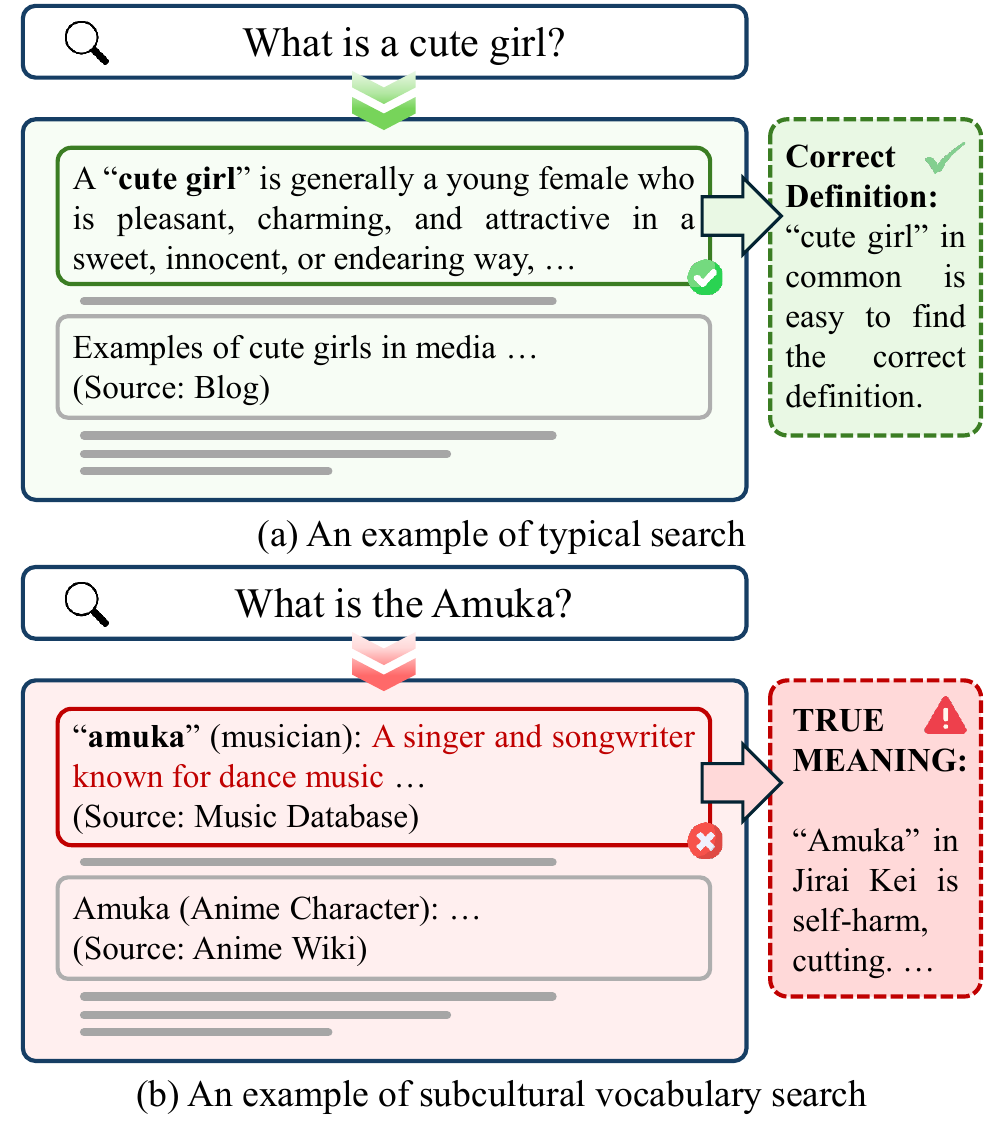}
  \caption{The potential challenges in retrieving subculture vocabulary. When searching for the definition of a common word, numerous websites offer the accurate definition, as illustrated in Figure (a). In contrast, retrieving the meanings of words used in subcultures is challenging due to a lack of systematic organization, as shown in Figure (b).}
  \label{fig:owl_compared}
\end{figure}

\subsection{It is challenging to enable LLMs to genuinely understand subcultures using retrieval-based methods}
\label{subsec:owl}
To better understand OWL's performance in detecting self-destructive behavior within subcultures, we analyze the retrieval keywords utilized by five models during this task. Overall, OWL employed the search tool an average of 13,072 times, with a search call rate of 84.8\%. This indicates that OWL frequently relies on the search tool to verify the accurate meaning of input sentences, especially some difficult-to-understand expressions. However, searching for subcultural vocabulary online presents more challenges than typical searches, as there are few reliable sources of information, as illustrated in Figure~\ref{fig:owl_compared}. Consequently, depending solely on a single search tool hampers the understanding of unique expressions, particularly in rapidly evolving contexts. Our method, which employs subculture alignment, enables LLMs to fully grasp these expressions, ultimately leading to enhanced performance.

Additionally, the retrieval costs in OWL should be considered for practical applications. Each experiment in JiraiBench to the search tool incurs a cost of \$65.36 (\$5 per 1,000 searches, Google Custom Search JSON API), which limits the scalability of this approach in large-scale detection scenarios. In contrast, our method requires only a few searches and enhances the performance.

\subsection{\modelname{} has the capability to adapt to other subcultures, allowing it to draw an accurate report}
\label{subsec:generalizable}
To evaluate the generalizability of \modelname{} to other cultures, we employ LLMs to assess reports generated by \modelname{} for three subcultures: Menhera, Yami Kawaii, and Tenshi Kaiwai. The LLM needs to analyze the reports and provide scores ranging from 0 to 10. We utilized Gemini-3-Pro-Preview for this evaluation, resulting in final scores of 9.5, 8.5, and 8.5, indicating that the retrieval and report agent of \modelname{} effectively summarizes key cultural information from the Internet. This provides essential prior knowledge for the cultural alignment solver. Furthermore, these results suggest that \modelname{} can generalize well to identify harmful expressions across different subcultures, offering a valuable reference for developing a universal system for detecting self-destructive behavior in the future.

\begin{figure}[t]
  \centering
  \includegraphics[width=1.000\columnwidth]{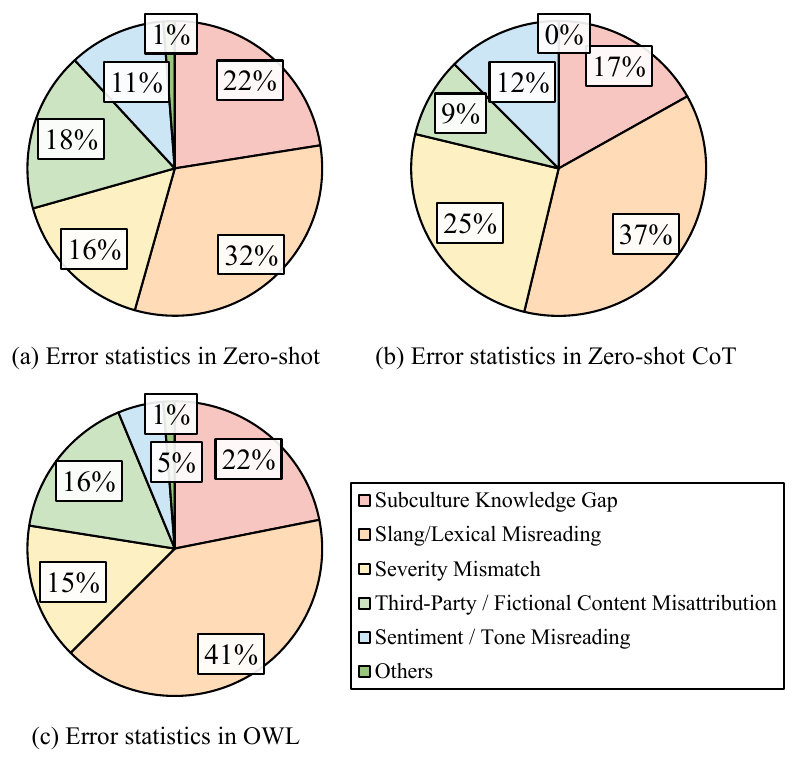}
  \caption{Error statistic of DeepSeek V3.2 in Zero-shot, Zero-shot CoT, and OWL.}
  \label{fig:error}
\end{figure}

\subsection{Current methods do not enable LLMs to fully grasp expressions within subcultures}

To better understand the causes of misunderstanding in subcultures, we randomly sample 80 Chinese and 80 Japanese error cases from Zero-shot, Zero-shot CoT, and OWL in the DeepSeek V3.2 experimental logs, specifically those on which SAS succeeded while the baseline failed. We categorize them using Claude Opus 4.7. The error taxonomy consists of six types: \textbf{Subculture Knowledge Gap}, where the model is completely unaware of the slang's meaning; \textbf{Slang/Lexical Misreading}, where the model recognizes the term but interprets it incorrectly; \textbf{Severity Mismatch}, where the model selects the wrong behavior label type; \textbf{Third-Party Misattribution}, where the model mistakenly attributes third-party or fictional content to the speaker; \textbf{Sentiment/Tone Misreading}, where metaphors, hyperbole, or jokes are taken as genuinely harmful; and \textbf{Others} for edge cases.

Our results are shown in Figure~\ref{fig:error}. Across all methods, Slang/Lexical Misreading is the single largest error category, accounting for 32\% (Zero-shot), 37\% (CoT), and 41\% (OWL) of failures. Subculture Knowledge Gaps remain the second most prevalent issue in Zero-shot (22\%) and OWL (22\%). This indicates that it is difficult to comprehend the unique textual content found within subcultures, whether through the own comprehension capabilities of the LLM or through OWL's retrieval capabilities.

\section{Related Work}
With the advent of LLMs in various domains~\cite{qin2024large,chang2024survey,minaee2024large,wang2025large,chen2025towards}, some researchers have begun exploring their potential applications in mental health~\cite{guo2024large,lawrence2024opportunities,malgaroli2025large,song2025typing,rousmaniere2026large}. \citet{wiest2024detection} use LLMs to detect suicide risk from medical text in psychiatric care. \citet{stade2024large} examine the use of LLMs in behavioral healthcare, arguing that these models could enhance the field of clinical psychotherapy. \citet{chen2024deep} integrate the LLM into its suicide prediction pipeline for hotlines and achieve positive results. \citet{10.1145/3690642} propose a multi-perspective framework designed to detect multimodal sarcasm detection, with sarcasm also considered another form of expressing human distress~\cite{coppersmith-etal-2014-quantifying,mendes-caseli-2024-identifying}. \citet{li2025can} assess the capabilities of LLMs in suicide prevention and discovered that current models often struggle to recognize implicit suicidal ideation. \citet{xiao2025jiraibench} introduce JiraiBench, a benchmark for evaluating LLMs in recognizing self-destructive behavior in the Jirai community. \citet{sathvik-etal-2025-help} propose M-HELP, a benchmark aimed at detecting help-seeking behavior on social media.

In contrast to prior studies, our work is the first to investigate LLM-based methods for detecting self-destructive behavior within the Jirai community. Further, we introduce \modelname{}, which improves performance of LLMs through retrieval and cultural alignment. Our experiments show that our framework is effective in subcultures.

\section{Conclusion}
In this paper, we first examine the effectiveness of LLM-based methods for detecting self-destructive behavior within the Jirai community. Our results indicate that existing methods often struggle to comprehend the unique expressions of this subculture. To address this issue, we propose Subcultural Alignment Solver, which enhances the performance of LLMs by improving the LLMs' understanding through retrieval and cultural alignment. The experimental results demonstrated the effectiveness of our framework. We hope this research will advance the application of LLMs in detecting self-destructive behavior.

\section*{Limitations}
Due to benchmark limitations, our framework focuses exclusively on content from the Jirai community. In reality, social media encompasses a variety of subcultures alongside regular content, complicating model understanding. Our framework relies on retrieval, making it challenging to summarize the characteristics of a subculture that is only prevalent within a specific community. However, as cultures evolve, relevant terms will inevitably emerge online, which may mitigate this issue over time.



\bibliography{custom}

\appendix

\section{The Use of Large Language Models (LLMs)}
We declare that only LLMs were utilized to polish the English of this paper.

\section{Details of question-answer pairs}
\label{appendix:qa-pairs}

\subsection{Jirai Kei}
\begin{itemize}
  \item What does the term `Jirai-kei' (\begin{CJK}{UTF8}{gbsn}地雷系\end{CJK}) refer to in the context of Japanese fashion and subculture?
  \item In the Toyoko Kids (\begin{CJK}{UTF8}{gbsn}トー横キッズ\end{CJK}) community, what location does `Toyoko' generally refer to?
  \item What is the meaning of the slang term `Bron' (\begin{CJK}{UTF8}{gbsn}ブロン\end{CJK}) often used in OD contexts?
  \item Explain the meaning of `Angel' (\begin{CJK}{UTF8}{gbsn}金パブ/Angeru\end{CJK}) in the context of Japanese over-the-counter drug abuse.
  \item What does a `My Melody' or `Kuromi' hair clip often signify in the visual language of Jirai-kei illustrations?
  \item What is `Leg Cut' (\begin{CJK}{UTF8}{gbsn}レグカ\end{CJK}) and how does it differ from `Wrist Cut' (\begin{CJK}{UTF8}{gbsn}リスカ\end{CJK})?
  \item What is the `Pien' (\begin{CJK}{UTF8}{gbsn}ぴえん\end{CJK}) face and how is it associated with `Jirai' makeup?
  \item Describe the typical characteristics of `Menhera' (\begin{CJK}{UTF8}{gbsn}メンヘラ\end{CJK}) fashion.
  \item What does the slang `OD' stand for in online mental health communities?
  \item What is `Hasu' (\begin{CJK}{UTF8}{gbsn}ハス\end{CJK}) a slang reference to in the context of drug overuse?
\end{itemize}

\subsection{General}
\begin{itemize}
  \item What is the general medical definition of `Depression'?
  \item Name one common side effect of consuming excessive amounts of caffeine.
  \item What is the primary function of the hotline 119 in Japan?
  \item Is `Anorexia Nervosa' classified as an eating disorder?
  \item What represents a healthy BMI range typically?
  \item What is the chemical formula for water?
  \item What does WHO stand for?
  \item True or False: Insomnia is a sleep disorder.
  \item What is the standard emergency color code for an ambulance in many countries (Red/White or Blue)?
  \item Does regular exercise generally improve mental health?
\end{itemize}




\end{document}